\documentclass[letterpaper, 10 pt, conference]{ieeeconf}  % Comment this line out if you need a4paper           

\IEEEoverridecommandlockouts  % This command is only needed if 
    
\usepackage{amsmath,amsfonts,amssymb}
\usepackage{algorithm}
\usepackage{float}
\newfloat{algorithm}{t}{lop}
\usepackage{xcolor}
\usepackage{algorithmic}
\usepackage{array}
\usepackage[caption=false,font=normalsize,labelfont=sf,textfont=sf]{subfig}
\usepackage{textcomp}
\usepackage{stfloats}
\usepackage{url}
\usepackage{verbatim}
\usepackage{graphicx}
\usepackage{cite}
\usepackage{booktabs}
\usepackage{todonotes}

\definecolor{highlander_blue}{RGB}{0,61,165}
\definecolor{olive}{RGB}{128,128,0}

\makeatletter
\newcommand\fs@spaceruled{\def\@fs@cfont{\bfseries}\let\@fs@capt\floatc@ruled
  \def\@fs@pre{\vspace{0.5\baselineskip}\hrule height.8pt depth0pt \kern2pt}%
  \def\@fs@post{\kern2pt\hrule\relax}%
  \def\@fs@mid{\kern2pt\hrule\kern2pt}%
  \let\@fs@iftopcapt\iftrue}
\makeatother

\begin{document}

\title{Language-guided Robust Navigation for Mobile Robots \\in Dynamically-changing Environments}

\author{Cody Simons, Zhichao Liu, Brandon Marcus, Amit K. Roy-Chowdhury, and Konstantinos Karydis
        % <-this % stops a space
        \thanks{The authors are with the Department of Electrical and Computer Engineering, University of California, Riverside, 900 University Avenue, Riverside, CA 92521, USA. Email: \{csimo005, zliu157, bmarc018, amitrc, karydis\}@ucr.edu. %}
        %\thanks{
        We gratefully acknowledge the support of NSF \# IIS-1901379, \# CNS-2312395 and ONR \# N00014-18-1-2252, \# N00014-19-1-2264. Any opinions, findings, and conclusions or recommendations expressed in this material are those of the authors and do not necessarily reflect the views of the funding agencies.}% <-this % stops a space%
        }

% The paper headers
%\markboth{IEEE ROBOTICS AND AUTOMATION LETTERS, VOL. \#, NO. \#, MONTH YEAR}%
%{Cody Simons, Brandon Marcus, Zhichao Liu, Konstantinos Karydis, Amit K. Roy-Chowdhury: Robotic Navigation with Natural Language Feedback}

%\IEEEpubid{0000--0000/00\$00.00~\copyright~2021 IEEE}
% Remember, if you use this you must call \IEEEpubidadjcol in the second
% column for its text to clear the IEEEpubid mark.

\maketitle

%%%%%%%%% ABSTRACT
\begin{abstract}
%There has been an explosion in robotic applications built on top of wheeled robots performing autonomous navigation. Leveraging prior maps greatly simplifies autonomous navigation, but real-world environments are rarely static. The prior map will diverge from reality as obstacles are added, removed, and rearranged. Handling these divergences automatically may result in sub-optimal solutions and, in the case of human-designed plans, violate the intent of the original plan. In these cases, it would be better to query a person for instructions on how to proceed. 
In this paper, we develop an embodied AI system for human-in-the-loop navigation with a wheeled mobile robot. We propose a direct yet effective method of monitoring the robot's current plan to detect changes in the environment that impact the intended trajectory of the robot significantly and then query a human for feedback. We also develop a means to parse human feedback expressed in natural language into local navigation waypoints and integrate it into a global planning system, by leveraging a map of semantic features and an aligned obstacle map.  Extensive testing in simulation and physical hardware experiments with a resource-constrained wheeled robot tasked to navigate in a real-world environment validate the efficacy and robustness of our method. This work can support applications like precision agriculture and construction, where persistent monitoring of the environment provides a human with information about the environment state.
\end{abstract}

%Since our method only uses machine learning to generate waypoints, we can still use conventional control and planning methods to navigate the environment. 

%\begin{IEEEkeywords}
%Visual Language Navigation, Vision-Based Navigation, AI-Enabled Robotics, Environment Monitoring and Management
%\end{IEEEkeywords}

%%%%%%%%% BODY TEXT
\section{Introduction}
%%%%%%%%%%%%%%%%%%%%%%%%%%%%%%%%%%%%%%%%%%%%%%%%%%%%%%%%%%%%%%%%%%%%%%%%%%%%%%%
%%%                 Para 1: General Mobile Robot Navigation                 %%%
%%%%%%%%%%%%%%%%%%%%%%%%%%%%%%%%%%%%%%%%%%%%%%%%%%%%%%%%%%%%%%%%%%%%%%%%%%%%%%%
%
% - Highlight mobile robot navigation in real-world applications
% - Center on environment monitoring and management
% - Show several examples of actual field deployment works with citations 
%       - Look in the Journal of Field Robotics
% - Convince reader this is rapidly becoming widespread
%
%%%%%%%%%%%%%%%%%%%%%%%%%%%%%%%%%%%%%%%%%%%%%%%%%%%%%%%%%%%%%%%%%%%%%%%%%%%%%%%
There has been a surge in the deployment of wheeled mobile robots in real-world settings across diverse applications. 
Sample applications include phenotyping~\cite{yang2024cropphenotyping} and leaf sampling~\cite{dechemi2023robotic} in precision agriculture, wall plastering~\cite{liu2024puttybot} in intelligent construction systems, persistent monitoring and inspection~\cite{wang2022persistentsurveillance,bird2022nucleardecommissioning}, and support services like cart collection~\cite{xie2024multitrolley}, to name a few. 
Central to a robot's successful deployment in the field is the ability to navigate through an environment autonomously. 

%To address this issue, we propose a method by which mobile robots navigating an environment may detect environmental changes, request feedback from a person, and process natural language to modify its behavior.

%%%%%%%%%%%%%%%%%%%%%%%%%%%%%%%%%%%%%%%%%%%%%%%%%%%%%%%%%%%%%%%%%%%%%%%%%%%%%%%
%%%                 Para 2: Runtime Uncertainties Occur in Practice         %%%
%%%%%%%%%%%%%%%%%%%%%%%%%%%%%%%%%%%%%%%%%%%%%%%%%%%%%%%%%%%%%%%%%%%%%%%%%%%%%%%
%
% - Robots work well in static environments
% - The real world is dynamic and uncertain
% - Highlight common types of uncertainty, especially those relevant to our work
%
%%%%%%%%%%%%%%%%%%%%%%%%%%%%%%%%%%%%%%%%%%%%%%%%%%%%%%%%%%%%%%%%%%%%%%%%%%%%%%%
%Autonomous navigation in static, cleanly observed environments is a largely solved problem, however, such environments only exist in simulation or highly controlled lab settings. 
Importantly, autonomous navigation in real-world settings is often subject to runtime uncertainties which cannot be modeled a-priori. 
A major source of uncertainty relates to the dynamics of the robot and the robot-environment interactions~\cite{karydis2015probabilistically}. 
This source of uncertainty, which is beyond the scope of this paper, can often be addressed via runtime robot learning~\cite{liu2024model,shi2024koopman}. 
A second major source of uncertainty, which this paper focuses on, concerns dynamic changes in the environment~\cite{wang2021navigation}.  
These can include moving obstacles (including humans)~\cite{fan2020uncertainty}, changes in ingress/egress points~\cite{nieuwenhuisen2010doors}, as well as variations in the semantic information often used for navigation. 
Such changes may occur in practice even when an overall environment map is known~\cite{pomerleau2014mapmaintenance}.%\todo{lines 33 and 34 need refs of works that have focused on each raised change in the env.} 
As a result, otherwise fine-tuned methods developed based on either classical SLAM tools (e.g.,~\cite{tian2022kimera}) or learning-based methods (e.g.,~\cite{han2020cooperative}), may still underperform and/or require re-tuning in face of dynamic changes in the environment. 

An alternative route to address such challenges can be via the integration of a human in the loop. 
The motivating idea is that certain assessments that may completely break an autonomous system (i.e. make the underlying joint perception and planning decision-making framework intractable during runtime) can be intuitively understood by a human, who, in turn, could provide appropriate feedback to the robot so that it can continue its operation. 
An illustrative example is differentiating (and adjusting accordingly) between a temporary obstacle (e.g., a person walking by) and a permanent obstacle (e.g., a closed door) that may appear during runtime execution in the robot's planned trajectory. 
In the former case, the robot would not need to replan but merely adjust its velocity, while the latter case should invoke a complete replanning of the path and follow-on trajectory~\cite{lu2020motion}. 

A critical component to the success of this form of interactive navigation is for the robot to determine both when and how to ask for help from a human. 
The frequency of queries to the human should be balanced so that the benefits of the interaction can be meaningful~\cite{amershi2014power}. 
To this end, several methods have been proposed, mostly following a reinforcement learning (or closely related) paradigm (e.g.,~\cite{mandel2017add,celemin2019interactive,spencer2022expert,xie2022ask,singi2024decision}) to learn a human-in-the-loop policy. 
However, as autonomous robots become increasingly more integrative in terms of perception, planning, and control, it may be hard to create sufficiently rich human-in-the-loop datasets for training\cite{majumdar2020improving,hahn2021norl}.
Our approach addresses this challenge by continuously assessing (re-)planned trajectory lengths in a metric map and asking for help once a user-defined threshold is exceeded. 

While there are multiple different interfaces to ask for and receive help, this paper posits that natural language can offer an immediate and effective bi-directional means for human-robot interaction. 
Guidance in natural language can be less error-prone (e.g., as compared to tapping on a screen a visual feature of interest, a waypoint to go to, etc.) and requires limited human training. %\todo{can we find references to the above two points made?} 
However, transforming natural language into information that can be actionable by the robot can be challenging~\cite{krantz2021waypoint}.
The advent of foundational models, crucially including vision language models (VLMs)~\cite{zhang2024vision}, along with the important capability to link text to image descriptions~\cite{radford2021clip} has offered a way to resolve this challenge. 
Indeed, vision-language maps (VLMaps)~\cite{huang2023vlmaps} have demonstrated the ability to associate language description with visual appearance features in metric-semantic maps, enabling several recent language-guided navigation works~\cite{chang2023goat,huang2023vlmaps,gadre2023cows}. 
We also leverage the ability to fuse semantic and metric information afforded via VLMaps to parse human input and create a set of revised waypoints for the robot to follow. 

The overall goal of this work is to create an embodied AI system that fuses the fast execution and reliability of conventional planning and control methods with the strong generalization capabilities afforded by contemporary learning-based methods for human-in-the-loop (wheeled) mobile navigation. 
We assume that a map of the overall operating environment is provided a-priori, but the environment is subject to dynamic changes during robot deployment that must be addressed for the robot to complete its navigation tasks within that dynamic environment. 
The contribution of this work is twofold. 
\begin{itemize}
    \item We propose a direct yet effective method for detecting when the robot should ask a human operator for assistance based on trajectory deviation using metric map information.
    \item We develop a framework by which a natural language query can be translated into navigation waypoints, which are then integrated into the global planning system and map information available to the robot.
\end{itemize}
The efficacy of our human-in-the-loop mobile robot navigation framework is confirmed via both simulation and physical experiments on a resource-constrained wheeled mobile robot.

\section{Related Works}
\subsection{Semantic Mapping}
Semantic mapping aims to augment a metric map of an environment with information about the semantic class of different objects. 
This information can be in the form of dense annotations~\cite{salas2013slam++} or object-level mappings~\cite{mccormac2018fusion++}.
Dense annotations typically aid in loop closure in SLAM methods~\cite{salas2013slam++}, but may also be used for higher-level scene understanding~\cite{rana2023sayplan}.
Object level mappings can use objects as high-level navigation landmarks~\cite{nicholson2019quadricSLAM} and may also be used to construct a scene graph for downstream applications~\cite{hughes2022hydra}.
Crucially, it has been possible to construct maps using open-vocabulary semantic features~\cite{huang2023vlmaps}. 
Using CLIP-based models~\cite{radford2021clip}, these maps can be queried with arbitrary text strings thus removing earlier limitations on the use of a pre-defined set of classes. 
Our work leverages dense semantic maps, focusing on how to integrate them with traditional planning methods directly.

\subsection{Language Guided Navigation}
The main premise in language-guided navigation is to enable a robot to move within an environment using natural language instructions from a human. 
Early works made use of scene graphs~\cite{majumdar2020improving,rana2023sayplan}, where each node of the graph contains some description of the local area, and navigation along the vertices is abstracted away. 
Navigating on the scene graph has been explored using both end-to-end learning~\cite{majumdar2020improving} and reinforcement learning techniques~\cite{majumdar2020improving}.

Recently there has been a focus on continuous navigation settings. 
In~\cite{irshad2021hierarchical,hong2021vlnbert} a reinforcement learning policy is learned to produce low-level motion primitives, while in~\cite{krantz2021waypoint} a learned waypoint picking policy is integrated with traditional controls. 
Large language models have also been leveraged to generate a navigation plan consisting of either action primitives~\cite{honerkamp2024momallm} or executable code~\cite{hu2024codebotler}. 
These methods do not assume access to pre-existing maps.  
However, in many practical cases (as in persistent monitoring applications for example), a map outline is available (or can be constructed first) but key elements within it may dynamically vary over different execution cycles. 
To this end, our work develops a decision-making module to query a human for help when significant changes in the map have occurred as well as a semantic map query module to parse human language feedback into waypoints in the associated metric map, and merges this embodied AI system with conventional planning and control methods for robust wheeled robot navigation in dynamically-changing environments. %\todo{Note from KK, no action needed. There is some repetition I have added here, maybe we can remove if space is needed.  But I feel it is important to keep highlighting to a reviewer what we are doing.}

\begin{figure*}[!t]
\vspace{6pt}
\centering
\includegraphics[trim={0cm, 1cm, 0cm, 1.5cm},clip,width=0.75\textwidth]{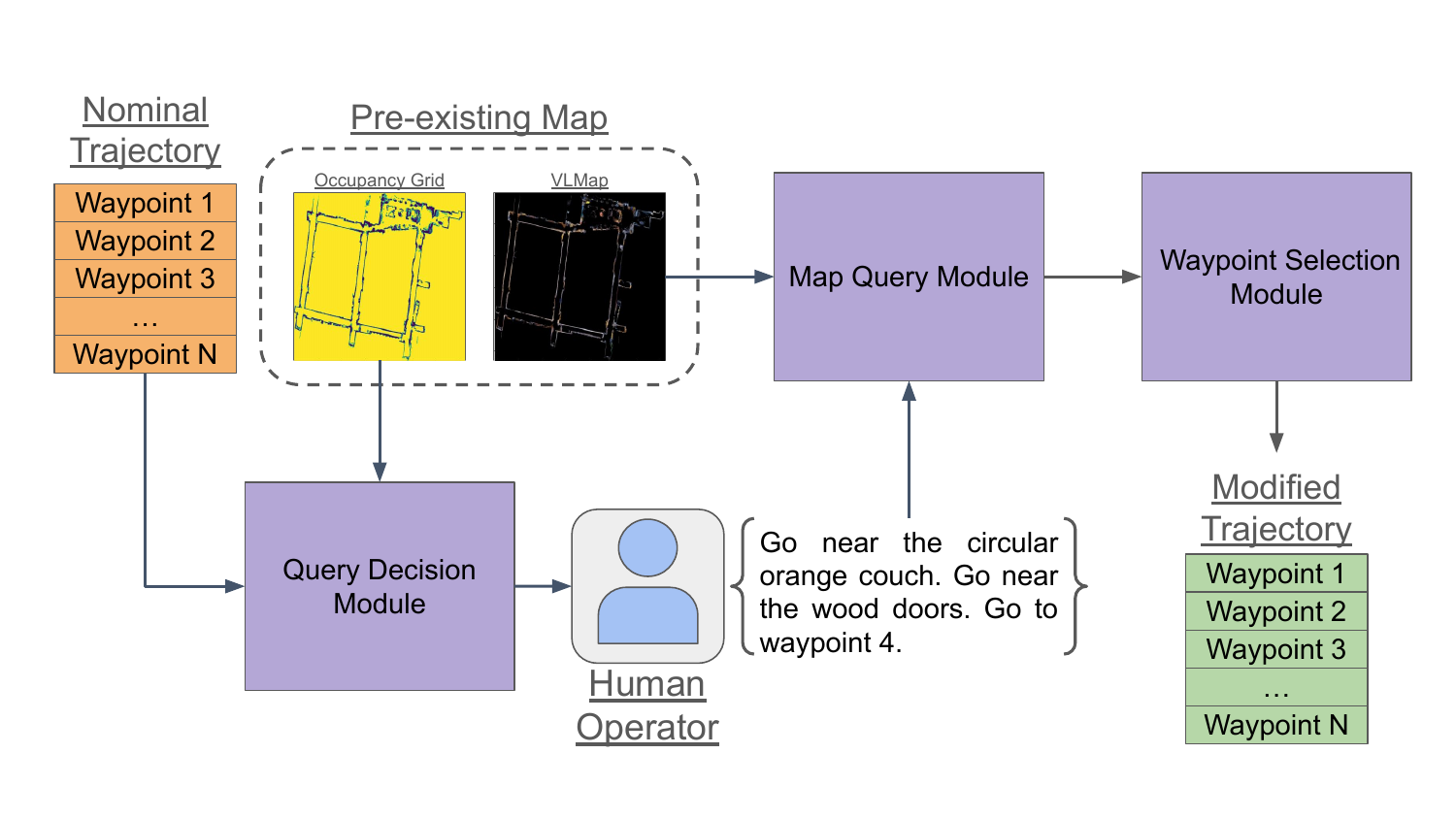}
\vspace{-6pt}
\caption{Overview of the human-in-the-loop navigation approach developed in this work. Given a nominal trajectory based on a prior (metric) map of the environment (along with the corresponding aligned semantic VLMap), total path deviations computed in real time invoke a query to the human once a threshold is exceeded. Then, human feedback is transformed from natural language into waypoints which are merged into the ongoing trajectory to generate the updated one for the robot to follow. This process repeats until a task is completed - herein to reach a specific pose on the map.}
%{In our method, a robot follows a nominal trajectory using a pre-existing map and off-the-shelf planning and navigation modules. When the \textit{Query Decision Module} decides that differences between the existing map and actual observations make the nominal trajectory unreasonable, a human operator is queried for natural language feedback. This feedback is used to query a visual-language map, using the \textit{Map Query Module} to find regions that match portions of the feedback. Finally, these matches are used by the \textit{Waypoint Selection Module} to create a modified trajectory to follow.}
\label{fig:method_overview}
\vspace{-12pt}
\end{figure*}

\section{Proposed Approach}\label{method}
Given an occupancy grid map, its corresponding VLMap, and a desired global navigation plan in terms of $N$ 2D waypoints, $\{(x_i, y_i, \theta_i)\}_{i=0}^N$, our method employs human feedback for online adjustment of local waypoints to circumvent local minima caused by dynamic changes in the environment. 
The overall approach is summarized in Fig.~\ref{fig:method_overview} and Alg.~\ref{algorithm}.
During operation, the \textit{Query Decision Module} monitors for differences between observations from the robot and the map, and, when the differences are significant, queries a human for aid. 
The human responds with natural language feedback, which is used by the \textit{Map Query Module} to query a semantic feature map and generate a set of navigation waypoint candidates. 
Then, the \textit{Waypoint Selection Module} builds a graph using the candidate waypoints, and the shortest path from the robot's current pose to the final waypoint becomes the new trajectory. 

\subsection{Query Decision Module}\label{method:query_decision}
The \textit{Query Decision Module} determines when to query a human for assistance by monitoring the current global navigation plan for significant changes from the nominal plan created based on the prior map (or an already updated plan in response to environment changes). 
We define as significant, changes that render current sensor observations incompatible with the map and may force the robot to get stuck in local minima, such as those caused by blocked passageways. 
This way, less crucial dynamic changes, such as a partially blocked hallway, or changes that do not affect navigation, such as appearance changes in surrounding objects, do not invoke a query.

%The global plan consists of densely packed points, which define a trajectory between navigational waypoints, which the robot will follow as it navigates through the environment. While navigating, 
The global plan is periodically recalculated (herein at a rate of $5$\;Hz). 
To detect significant changes, we compare the lengths of the new and previous plans. 
When their deviation exceeds a certain threshold, $\tau>0$, a query to the human is invoked.\;\footnote{~While there are other ways to invoke a query to the human, as described in the related literature, the total path deviation considered here affords direct computation (which is crucial for online adaptation as we do) and pairs well with the intended task of the robot, that is persistent navigation.} 
The threshold's value determines the degree of reliance on the human and can be selected based on the criticality of the application and robot dynamics. 
In this work, we empirically identified that setting $\tau=25\%$ can strike a balance in all tested cases (both in different simulated environments and in real world experiments).
%Otherwise, the robot continues with the current plan. 
%Typically, plan recalculations are small and decrease as the robot approaches its current goal, so the ratio is less than $1$. However, if an unexpected obstacle is blocking the robot, such as a closed door, this will result in a large increase in the length of the global plan. %The new plan will exceed the threshold, and the \textit{Query Decision Module} will request assistance from a human operator.

\subsection{Map Query Module}\label{method:map_query}
%Once feedback is received, it must be translated into a sequence of new navigation waypoints. 
Once a query is sent, the human is expected to provide feedback in the form of descriptive phrases corresponding to specific locations within the environment. 
The \textit{Map Query Module} processes each description and produces one or more candidate waypoints located in metric map areas that correspond to those descriptions. 

%To create the corresponding VLMap, 
Images from the environment were collected and used to build the corresponding VLMap, that is, a semantic feature map of contrastive vision-language features, as in~\cite{huang2023vlmaps}. 
The VLMap associates each cell location $(x,y)$ in the 2D occupancy grid map with a visual feature vector of length $M$, $f_{x,y}\in\mathbb{R}^M$. 
%, with $K$ the dimension of the feature vector. 
This map can be queried with arbitrary natural language descriptions, $d$, encoded using the CLIP~\cite{radford2021clip} text encoder $\mathcal{E}_{text}$, by taking the inner product to get a score $s_{x,y}=f_{x,y}^T\mathcal{E}_{text}(d)$ for how well each cell corresponds with text description, $t$. 
Since these scores are unit-less scalars, setting an arbitrary threshold for a positive match is ineffective and does not handle the varying degrees of affinity between positive matches, i.e. \textit{couch} and \textit{chair} both have a high score for most types of seating. 
Prior work~\cite{huang2023vlmaps} handled this problem by querying with multiple descriptions and assumed the highest score was correct. 
However, this requires that every location matches at least one of the queries and enforces querying for many background classes, thereby increasing computational complexity.

To address this shortcoming, we determine positive matches by examining the distribution of the query scores. 
A $k$-component Gaussian mixture model~\cite{bilmes1998gentle} is fit to the scores across the entire map, resulting in a set of means $\{\mu_i\}_{i=0}^k$ and a set of standard deviations $\{\sigma_i\}_{i=0}^k$. A score is considered a match if it belongs to the normal distribution with the highest mean, that is
\begin{equation}
\label{EQ: positive match}
\mathbf{1}(\textit{arg}\max_i \mathcal{N}(s_{x,y}; \mu_i, \sigma_i^2) = \textit{arg}\max_i \mu_i)\;,
\end{equation}
and all other possible cells are ignored. 
The output binary map may still be noisy. 
Applying basic filtering methods using morphological operations can result in a map that contains several contiguous regions and little noise. 

For each contiguous region, we construct a map of points within a fixed radius and known to be free space in the occupancy grid map. 
We calculate the clearance of each point to known obstacles %using a flood-fill~\cite{smith1979floodfill} algorithm 
and then sample a waypoint from the set of points with maximal clearance. 
This process is repeated for each matching region, giving a set of waypoints corresponding to that description. 
The entire procedure repeats for each description, producing a sequence of sets of candidate waypoints.

% \subsection{Waypoint Selection Module}\label{method:waypoint_selection}
% The binary map produced by the map query module is used by the waypoint selection module to produce waypoints, which are passed to the navigation \& planning modules. This is done by first sampling candidate waypoints from regions corresponding to different pieces of feedback and then constructing a graph from these waypoints. The floyd-warshall\cite{floyd1962allshortestpath} shortest path algorithm is used to pick a final set of waypoints to pass to the planning \& navigation modules.

% \subsubsection{Sampling Method}
% The filtered binary map is used to produce waypoints that correspond to different descriptions in the human response. Typically, a single description will match multiple areas in the map. Each contiguous region produces a single waypoint. If two areas are sufficiently close, they will be treated as a single region. 

% Using the occupancy grid from the map, a new region is created, which is near the region of interest, sub-one-meter distance, and known to be free space. For each point in this new region, the distance to the border is calculated. This can be done efficiently using the flood-fill algorithm\cite{smith1979floodfill}. A waypoint is sampled uniformly randomly from the points that are maximally far from the border. This process is repeated for every region of interest in a query map and for every query map produced by the human feedback to produce the complete set of candidate waypoints.

\floatstyle{spaceruled}
\restylefloat{algorithm}
\begin{algorithm}[!ht]
    \caption{Language-guided Waypoint Navigation}\label{algorithm}
    \begin{algorithmic}
        \REQUIRE Plan $=\{(x_i, y_i, \theta_i)\}_{i=0}^N$
        \STATE Begin navigating to $(x_0, y_0, \theta_0)$
        \STATE Get initial trajectory $p$ from robot
        \WHILE{$(x_N, y_N, \theta_N)$ is not reached}
            \STATE Wait for new trajectory $p'$ from robot
            \IF{QueryDecisionModule($p$, $p'$)}
                \STATE Request feedback from human and store in $feedback$
                \STATE Initialize Candidate Waypoint list $w$ with robot's current pose
                \FOR{each text description $d$ in the $feedback$}
                    \STATE Append waypoints MapQueryModule($d$) to $w$
                \ENDFOR

                \STATE WaypointSelectionModule$(w)\rightarrow$ Plan
                \STATE Begin navigating to $(x_0, y_0, \theta_0)$
                \STATE $p$ is the current trajectory
            \ELSE
                \STATE $p'\rightarrow p$
            \ENDIF
            \IF{Current goal reached}
                \STATE Begin navigating to $(x_{i+1}, y_{i+1}, \theta_{i+1})$
                \STATE Get initial trajectory $p$ from robot
            \ENDIF
        \ENDWHILE
    \end{algorithmic}
\end{algorithm}

\subsection{Waypoint Selection Module}\label{method:waypoint_selection}
The \textit{Waypoint Selection Module} selects a final sequence of waypoints from the candidate waypoints generated in the previous step. 
%To do so, %we construct a graph and employing classical planning methods. 
A directed graph is constructed where every candidate waypoint and the current robot pose are the vertices, and edges exist only between waypoints corresponding to sequential feedback pieces. 
%An example of such a construction is shown in Fig \ref{fig:graph_construction}. 
Edge weights are computed based on the estimated travel distance in the map. 
If two waypoints are too close, which may happen if two similar descriptions are given sequentially by the human, this connection can be pruned for computational expediency~\cite{lu2021deformation}.

Once the graph is created, the Floyd-Warshall algorithm~\cite{floyd1962allshortestpath} is used to find the shortest path between all waypoints. 
This is needed to handle cases when the final piece of feedback matches multiple locations.\;\footnote{~We note here that other online search-based algorithms are applicable in practice (e.g.,~\cite{liu2017search,lu2022online}), albeit they may incur additional computational cost if deployed to handle multiple terminal nodes - a detailed treatise of this topic is part of future work enabled by this current effort.}
%If we had used Dijkstras or a similar path planning algorithm in cases with multiple terminal nodes, the planning algorithm would need to be run multiple times. 
The shortest path between the robot's current pose and any waypoint linked to the last part of feedback becomes the new sequence of waypoints for the robot. %to pass through. 
Since waypoints are only connected if they correspond to sequential pieces of feedback, this new trajectory is guaranteed to have exactly one waypoint corresponding to each description in the human feedback.

\section{Experiments, Results, and Discussion}\label{experiments}
We tested our method extensively both in simulation and experimentally. %demonstrate our methods both in simulation and real-world experiments. 
In every setting, we considered multiple scenarios, including some with multiple feedback requests, and conducted multiple trials to evaluate the method's robustness. 
Further, in simulation, we performed an ablation study to better highlight the impact of the language processing method in the overall human-in-the-loop navigation framework.

\subsection{Simulation Setup and Results}\label{experiment:simulation_result}
We used a digital twin of Husarion RosBot2.0 Pro robot, equipped with an RGBD camera and a 2D LiDAR, in two different simulated environments in Gazebo. 
The first environment is the small house environment, released by AWS Robomaker, emulating a small residential house consisting of a bedroom, living room, kitchen, and dining room. 
The second environment is a factory floor, with shelving and other obstacles consistent with a factory setting. 
Three different routes were considered in each environment, and five trials were conducted for each route. 
For fairness, the same instructions were provided in each trial of a particular route, decided by a human operator, when the robot generated
a query. 
All simulations were run on a desktop PC with an Intel i9 CPU and Nvidia 3090ti GPU.

\begin{figure}[!t]
\vspace{6pt}
    \centering
    \subfloat[]{%[Gazebo Environment]{
        \reflectbox{
                \frame{\includegraphics[angle=180,width=0.85\linewidth]{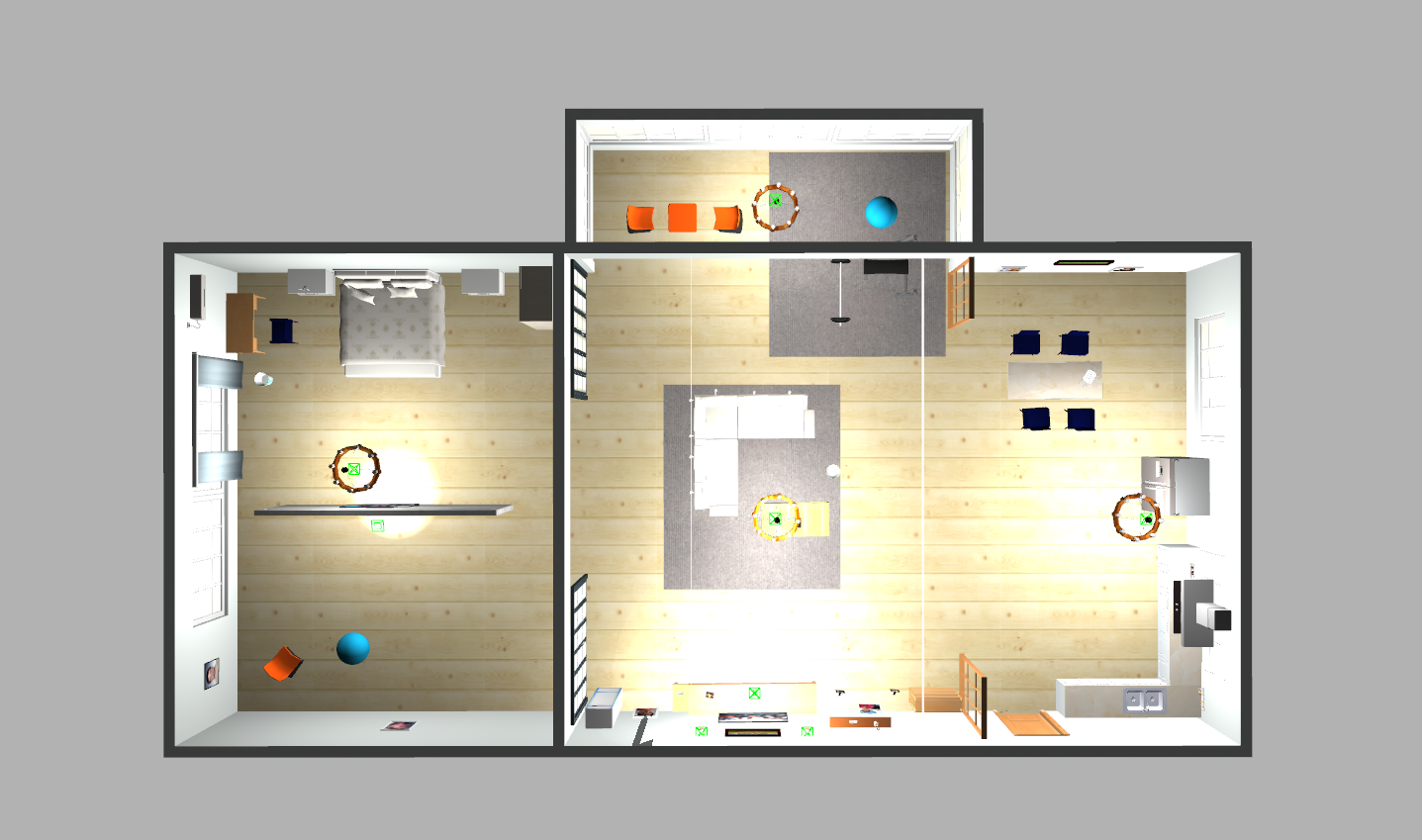}}
        }
    }%
    \\
    \subfloat[]{%[RGB Map]{
        \reflectbox{
            \includegraphics[angle=90,width=0.85\linewidth]{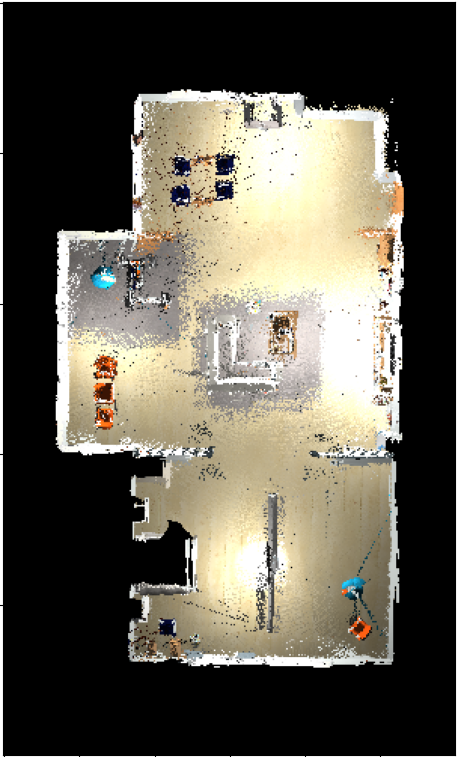}
        }
    }%
    \caption{(a) Top-down view of the small house environment rendered in Gazebo and (b) the corresponding RGB map created during our map construction process. Note that while the major features agree, there can still be inexact edges caused by sensor noise.}% which was not present in previous simulations.}%
\label{fig:small_house_map}
\end{figure}

To create the prior maps of each environment we manually operated the robot, employed the GMapping~\cite{grisetti2007gmapping} package in ROS to create the 2D occupancy grid, and used the pose estimated by GMapping along with collected RGBD information to construct the associated semantic feature map. %by a person familiar with the scene layout. 
%The obstacle map is built using the 
%The pose estimated by Gmapping is also used with collected RGBD images to build the semantic feature map. T
Pose estimation introduced some noise into the map construction process (Fig.~\ref{fig:small_house_map}). %for the small house environment. 
With the prior map available, we spawned additional obstacles into each environment as a means to introduce dynamic changes in the environment and trigger queries during execution. %\textit{Query Decision Method}.

\begin{table}[!t]
    \centering
    \caption{Simulation Results.}\label{tab:simulation_system_test}
    \vspace{-6pt}
    \begin{tabular}{ccccc}
        \toprule
        & \multicolumn{2}{c}{Small House} & \multicolumn{2}{c}{Factory}\\
        \cmidrule(lr){2-3}\cmidrule(lr){4-5} 
        & RMSE & SR & RMSE & SR\\
        \midrule
        Route 1 & $0.507\pm0.100$ & $1.00$ & $0.324\pm0.046$ & $1.00$\\
        Route 2 & $0.442\pm0.061$ & $1.00$ & $3.056\pm1.881$ & $1.00$\\
        Route 3 & $0.460\pm0.091$ & $1.00$ & $0.989\pm1.215$ & $0.80$\\
        Average & $0.470\pm0.090$ & $1.00$ & $1.456\pm1.739$ & $0.93$\\
        \bottomrule
    \end{tabular}
    \vspace{-12pt}
\end{table}

\begin{figure*}[!t]
    \centering
    \captionsetup[subfigure]{labelformat=empty}
    \subfloat[]{%[Small House]{
        \subfloat[(a)]{%[Route 1]{
            \frame{\includegraphics[trim=115 100 102 90, clip, width=0.3\textwidth]{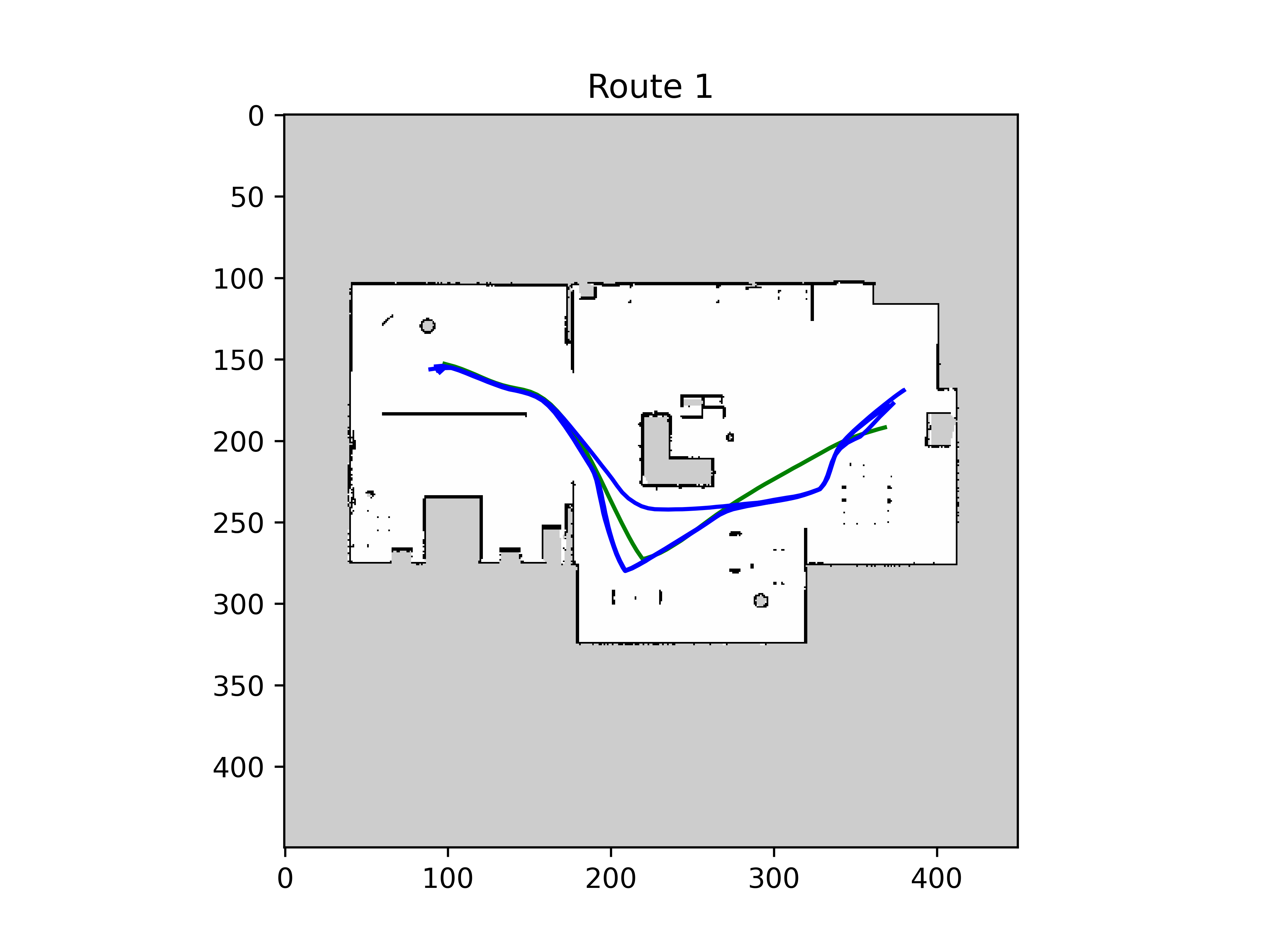}}
        }%
        \subfloat[(b)]{%[Route 2]{
            \frame{\includegraphics[trim=115 100 102 90, clip, width=0.3\textwidth]{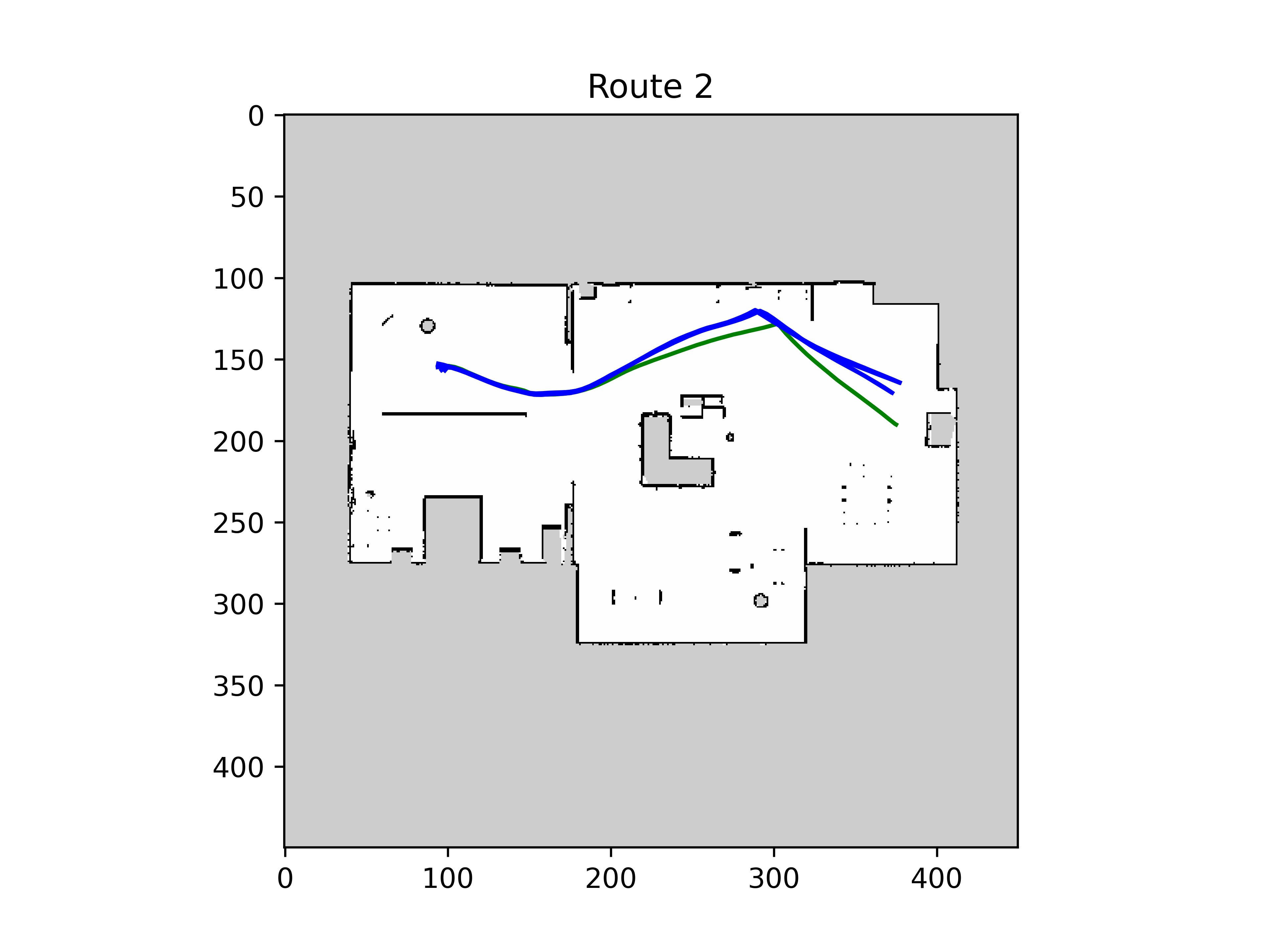}}
        }%
        \subfloat[(c)]{%[Route 3]{
            \frame{\includegraphics[trim=115 100 102 90, clip, width=0.3\textwidth]{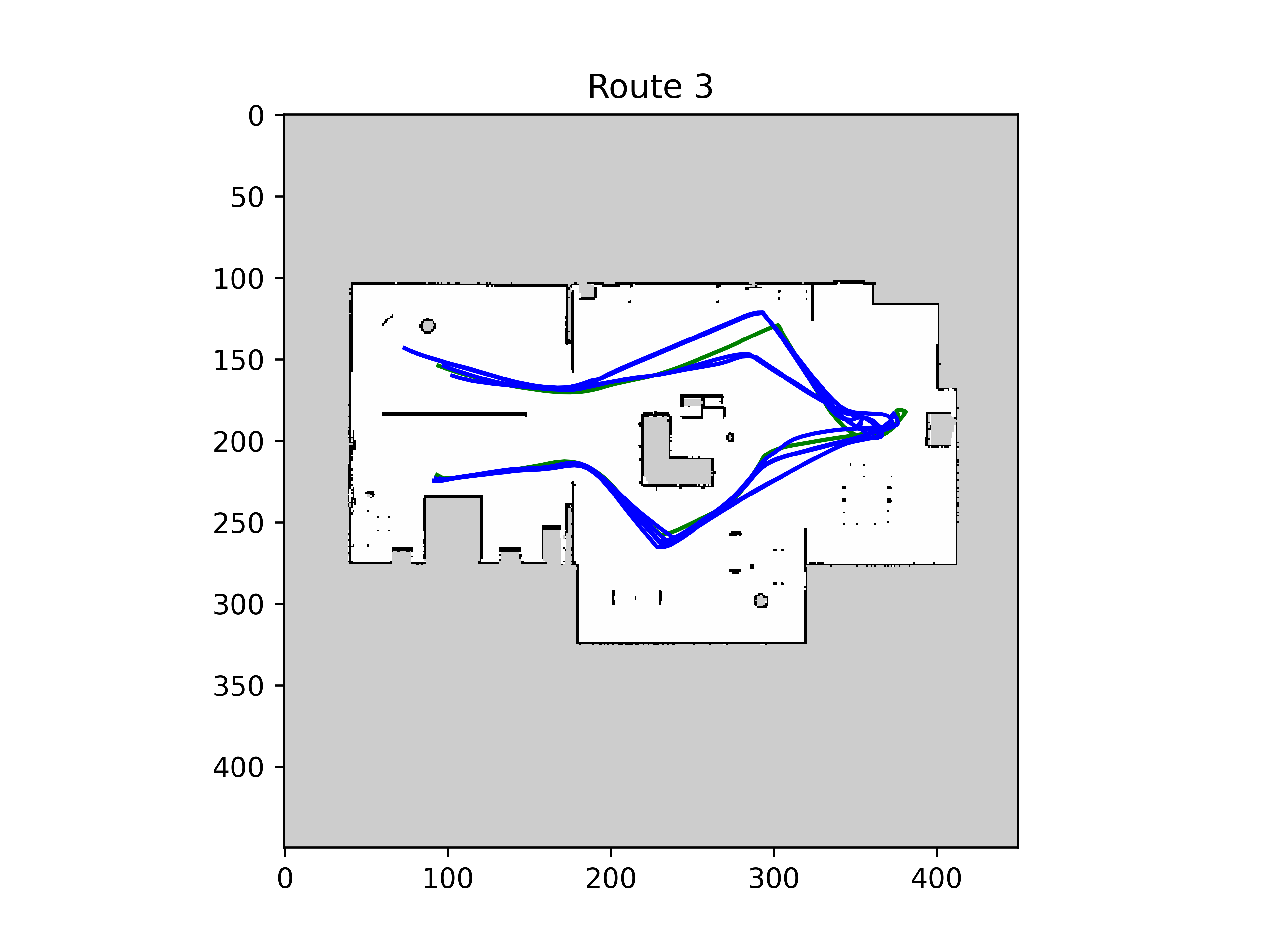}}
        }%
    }
    \vspace{-19pt}
    \subfloat[]{%[Factory]{
        \subfloat[(d)]{%[Route 1]{
            \frame{\includegraphics[trim=120 110 110 110, clip, width=0.3\textwidth]{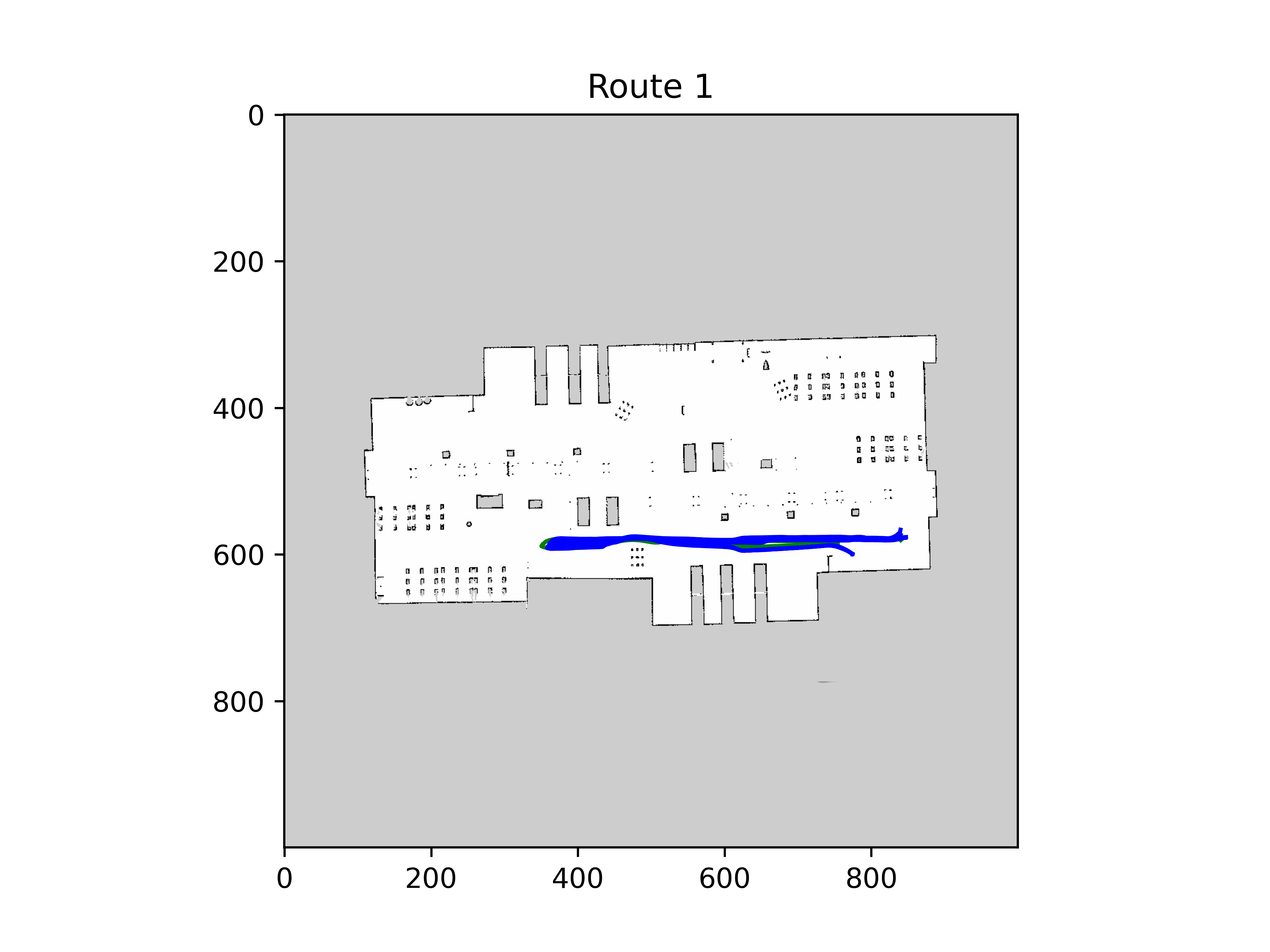}}
        }%
        \subfloat[(e)]{%[Route 2]{
            \frame{\includegraphics[trim=120 110 110 110, clip, width=0.3\textwidth]{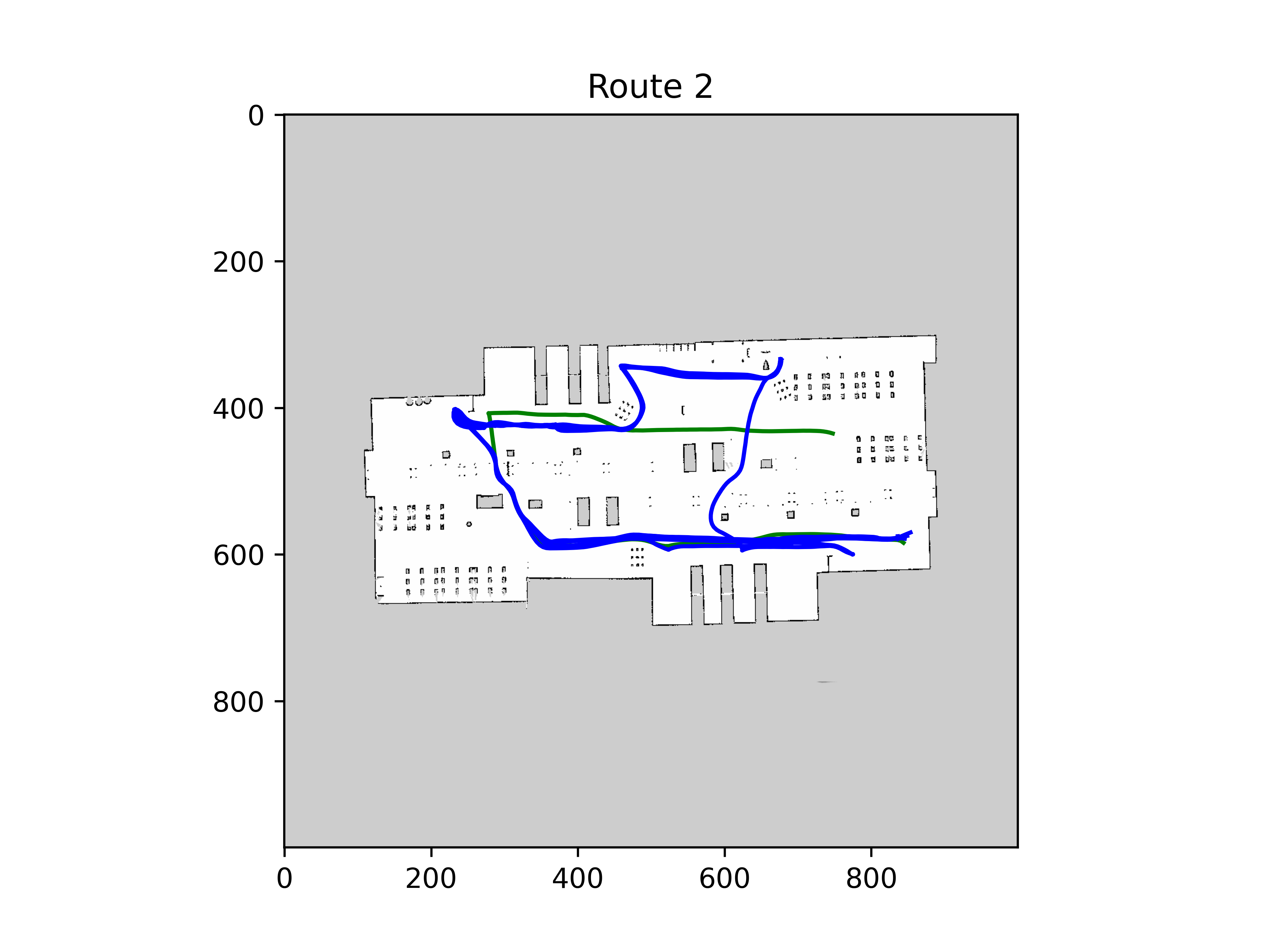}}
        }%
        \subfloat[(f)]{%[Route 3]{
            \frame{\includegraphics[trim=120 110 110 110, clip, width=0.3\textwidth]{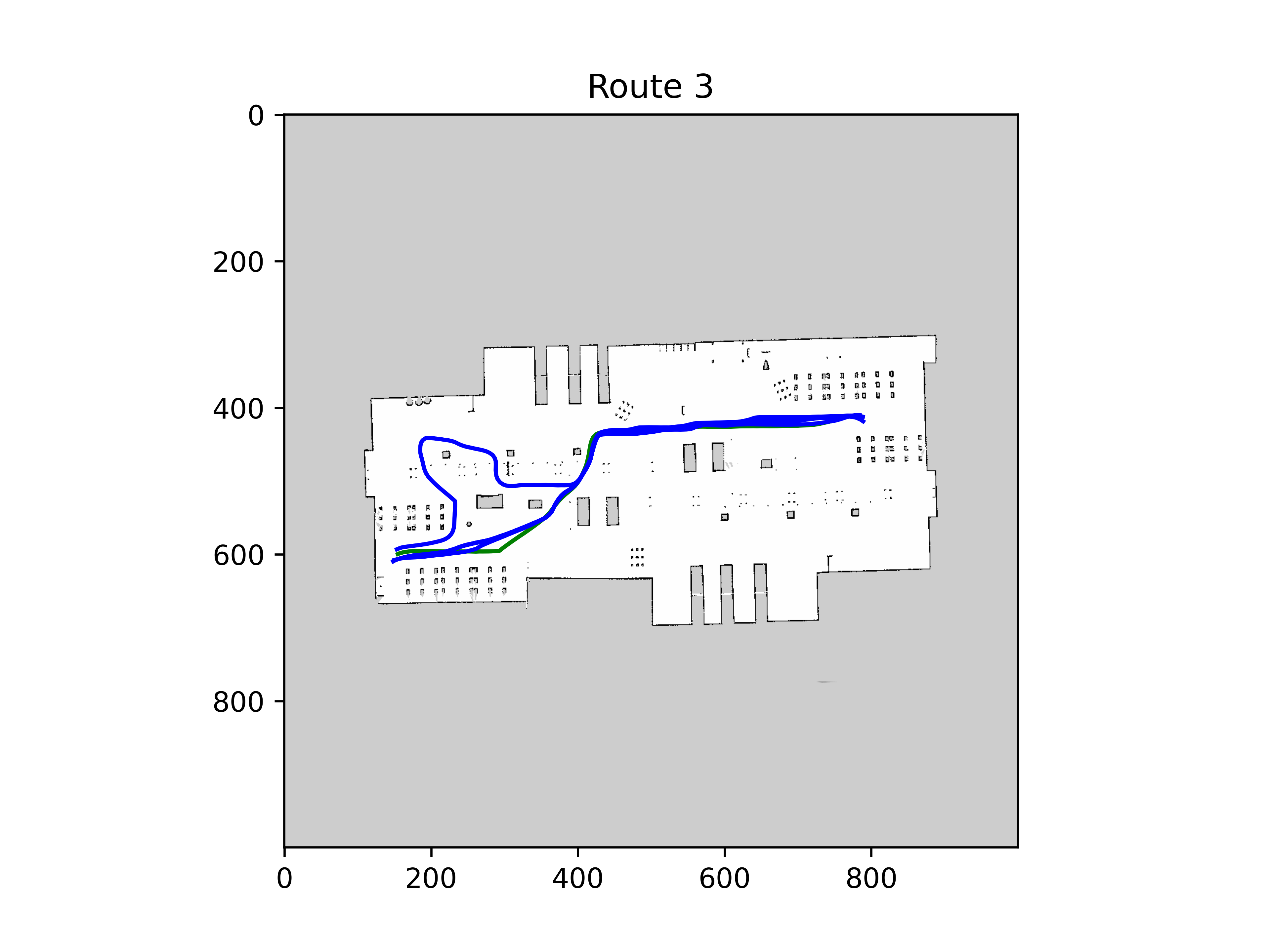}}
        }%
    }
    \vspace{-16pt}
    \caption{Trajectories generated by our method (blue) overlaid to the ground truth (green) in simulation testing. Panels (a)-(c) and (d)-(f) correspond to the small house and factory environments, respectively, for three distinctive routes.} %{Note that the cases with the routes with high RMSE include erroneous waypoints, which bring part of the trajectory away from the ground truth.}%
\label{fig:simulated_system_traj}
\vspace{-12pt}
\end{figure*}

Table~\ref{tab:simulation_system_test} contains the RMSE computed between the routes generated by our method and a ground truth trajectory manually created by a researcher, and the success rate (SR) for the \textit{Query Decision Module}. 
The ground truth trajectory was limited to the same number of waypoints and was given the same feedback to guide waypoint selection. 
Figure~\ref{fig:simulated_system_traj} depicts all trials for each considered case.

RMSE values are generally low in all cases, except one instance in the factory environment (Fig.~\ref{fig:simulated_system_traj}(e)). 
In that case, our method selected a feature that was compatible with the language instruction but was far from the ground-truth waypoint. 
Such large deviations (while still successful in reaching the goal) can be caused when multiple similar features may be available and compete against each other. 
This can be rectified by introducing follow-on queries (i.e. asking for clarification), and is part of ongoing work. 
Success rates are also high ($100\%$) in all but one of the tested cases where it drops to $80\%$ (Fig.~\ref{fig:simulated_system_traj}(f)). 
This case also occurs in the factory environment but at a different route than the case of higher RMSE. 
In this case, the changes in the global plan were too gradual to trigger a query, thus preventing crucial human feedback when needed to avoid local minima. 
This shows the need for human-in-the-loop navigation in dynamically-changing environments.

\begin{figure*}[!t]
    \centering
    \captionsetup[subfigure]{labelformat=empty}
    \subfloat[]{%[Small House]{
        \subfloat[(a)]{
            \frame{\includegraphics[trim=115 100 102 90, clip, width=0.3\textwidth]{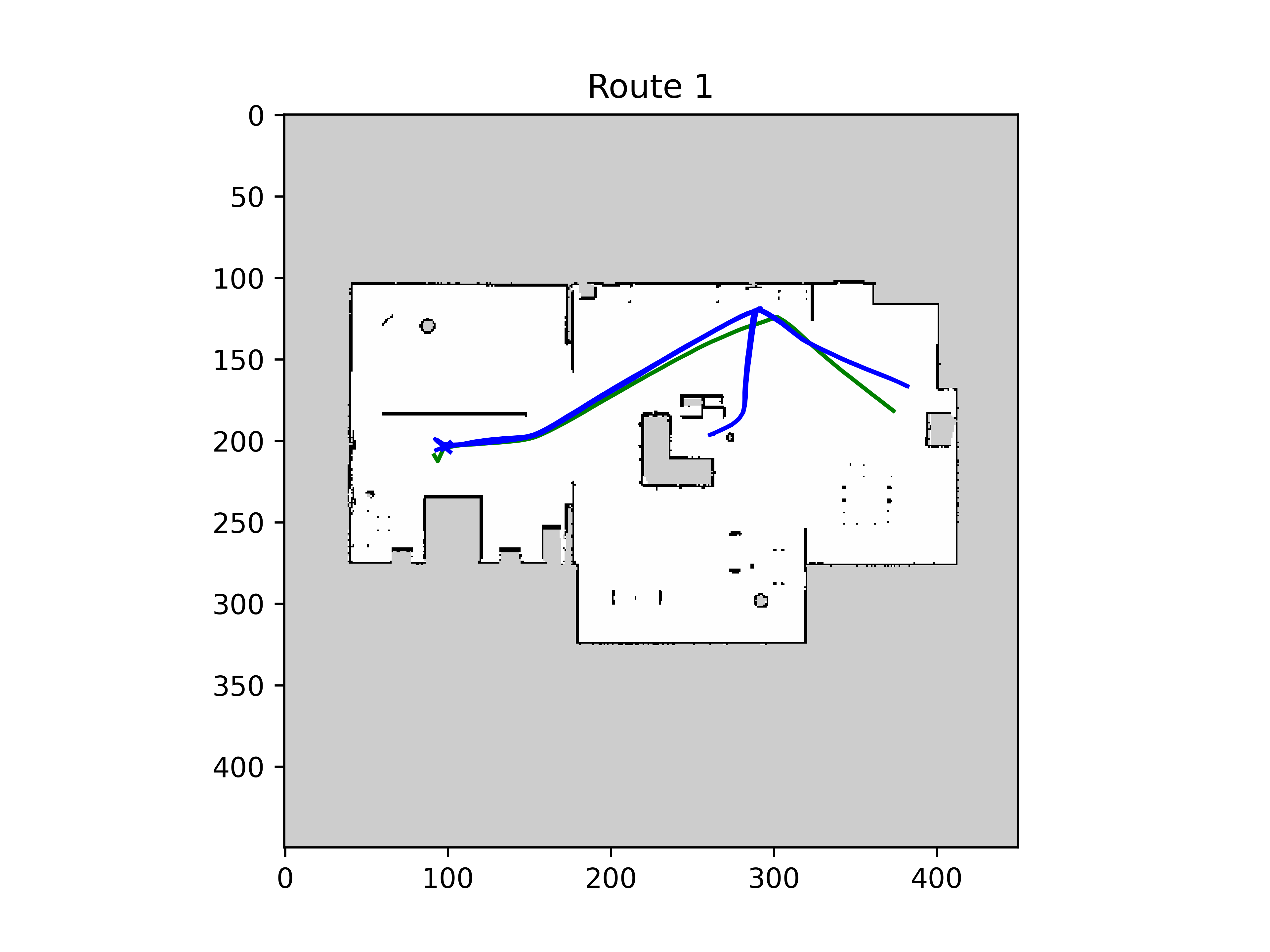}}
        }%
        \subfloat[(b)]{
            \frame{\includegraphics[trim=115 100 102 90, clip, width=0.3\textwidth]{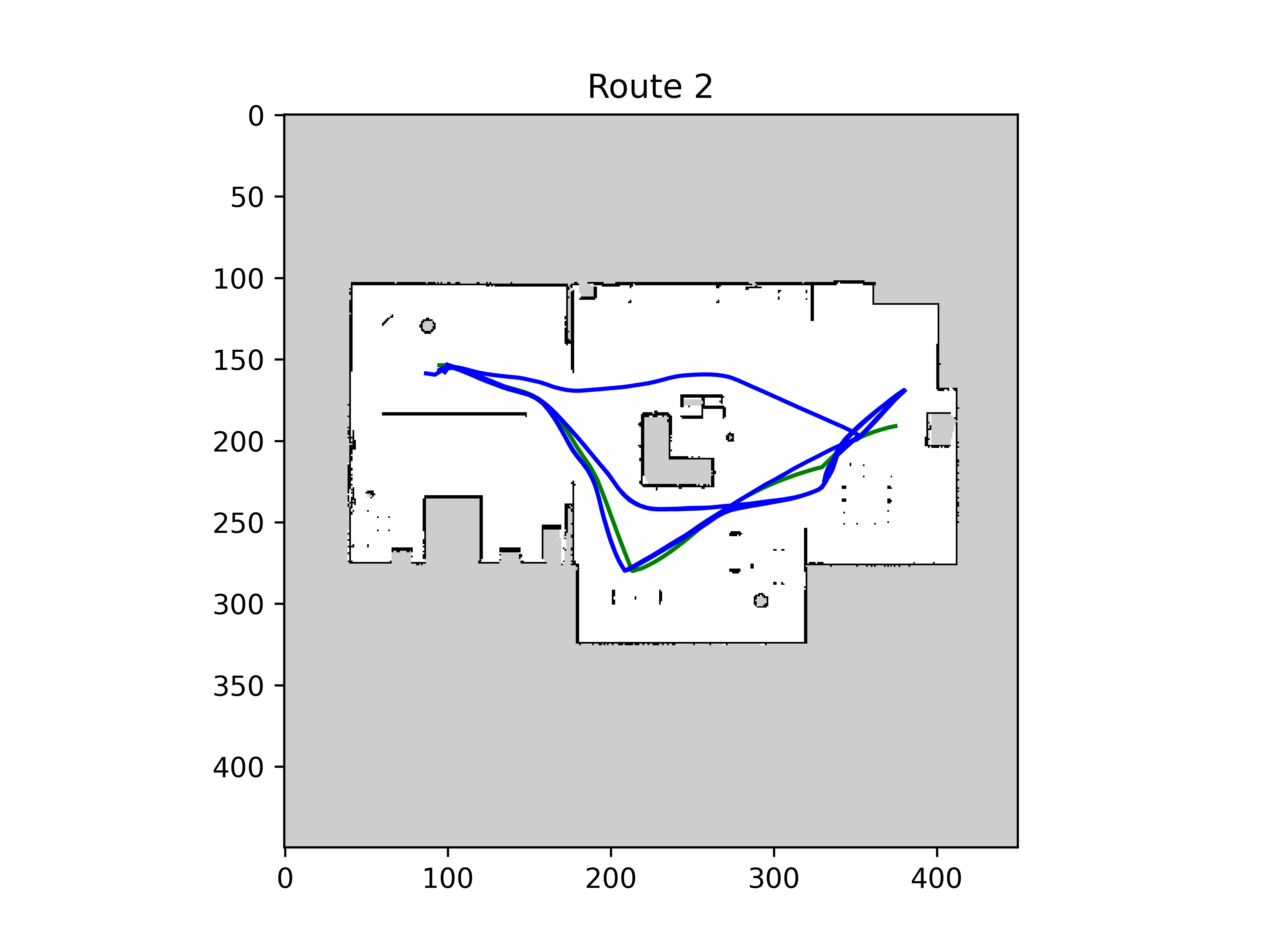}}
        }%
        \subfloat[(c)]{
            \frame{\includegraphics[trim=115 100 102 90, clip, width=0.3\textwidth]{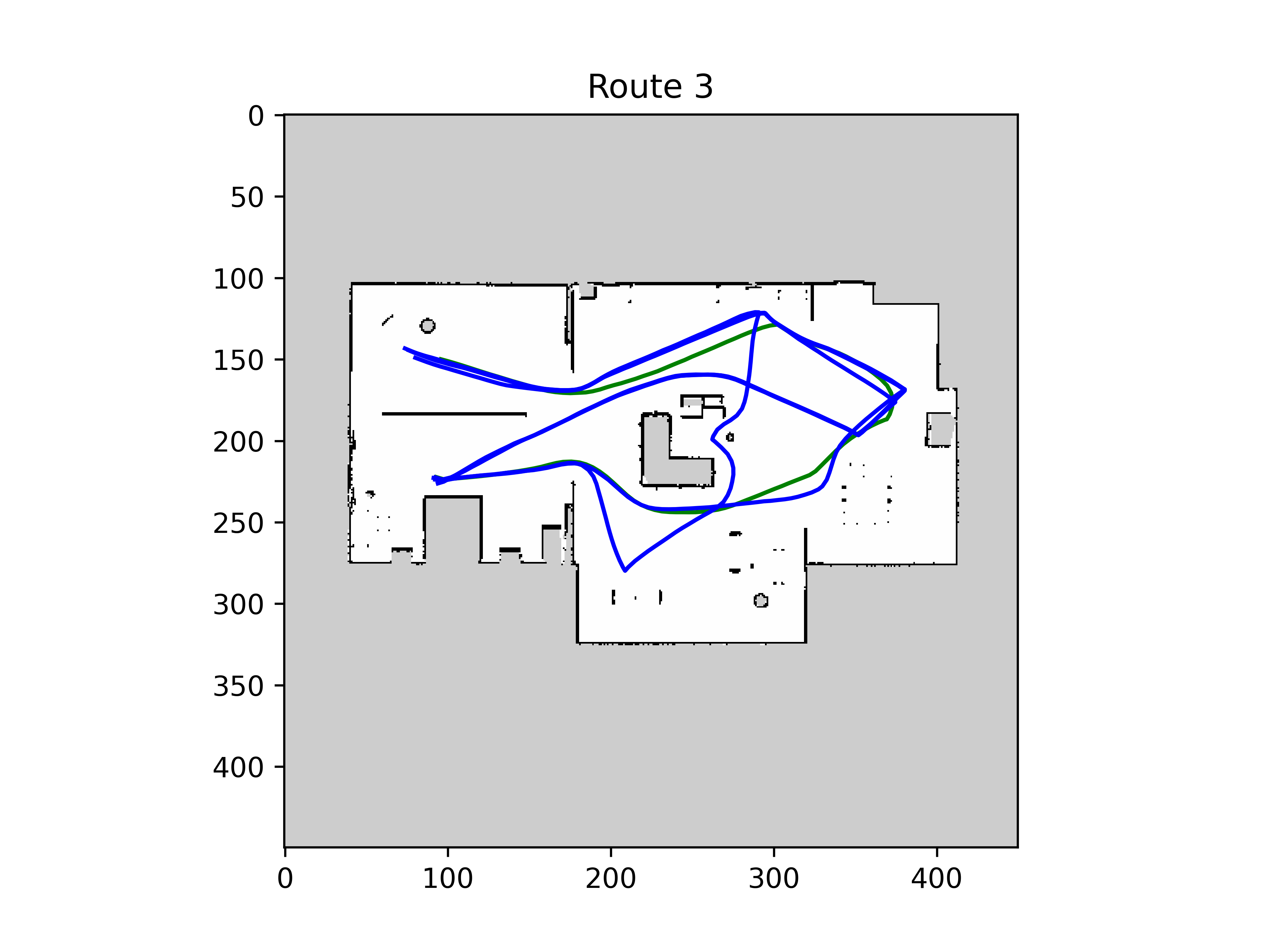}}
        }%
    }
    \vspace{-19pt}
    \subfloat[]{%[Factory]{
        \subfloat[(d)]{
            \frame{\includegraphics[trim=120 110 110 110, clip, width=0.3\textwidth]{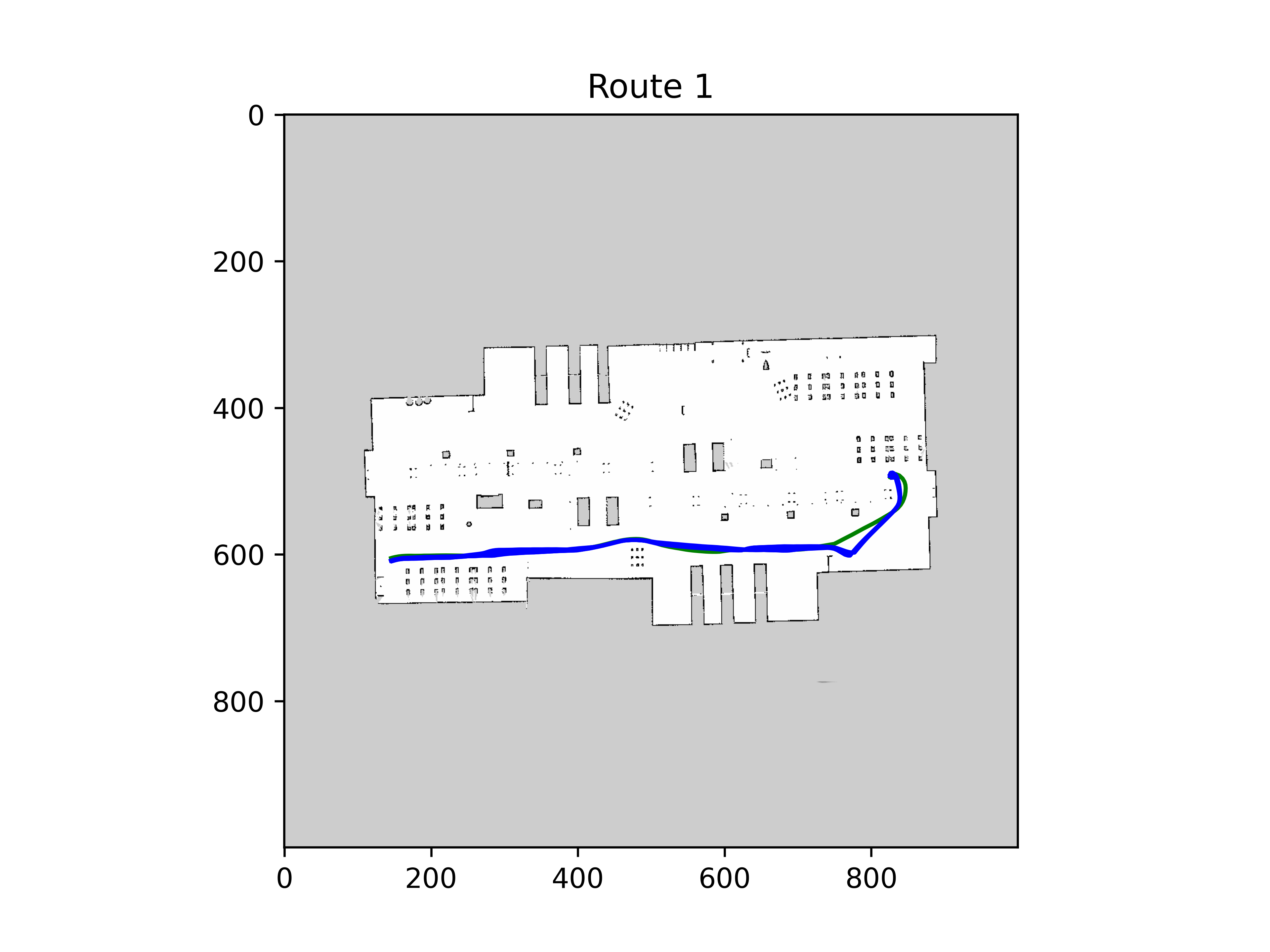}}
        }%
        \subfloat[(e)]{
            \frame{\includegraphics[trim=120 110 110 110, clip, width=0.3\textwidth]{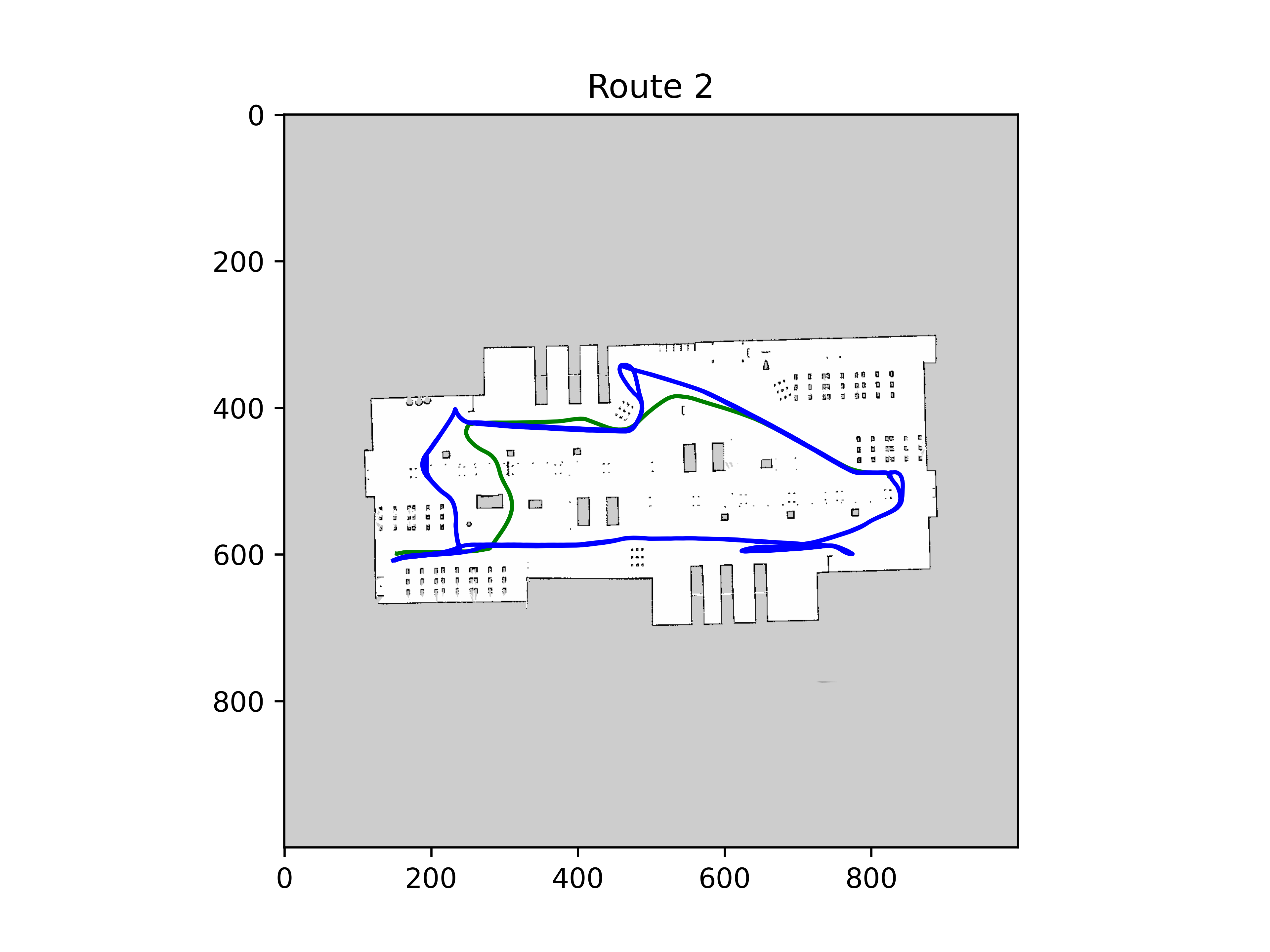}}
        }%
        \subfloat[(f)]{
            \frame{\includegraphics[trim=120 110 110 110, clip, width=0.3\textwidth]{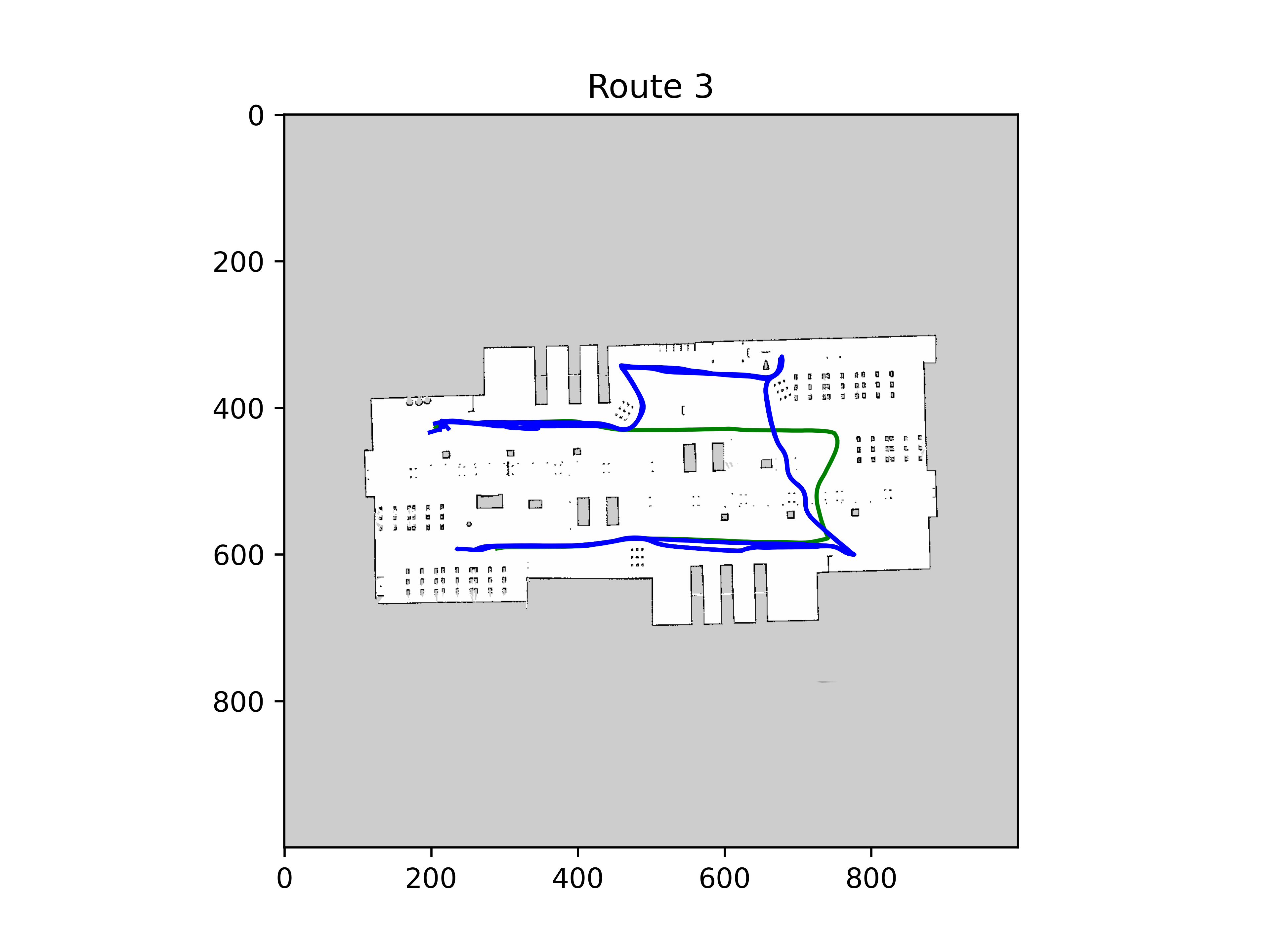}}
        }%
    }
    \vspace{-16pt}
    \caption{Trajectories generated by our method (blue) overlaid to the ground truth (green) in the language navigation ablation study (in simulation). Panels (a)-(c) and (d)-(f) correspond to the small house and factory environments, respectively, for three distinctive routes.} %\caption{Visualized results of our language navigation ablation study, with each trajectory of our method drawn in blue and the ground truth drawn in green. Note that the cases with the routes with high RMSE include erroneous waypoints, which bring part of the trajectory away from the ground truth.}%
\label{fig:simulated_lang_traj}
\vspace{-12pt}
\end{figure*}

\subsection{Language Navigation Ablation Study}\label{experiment:lang_nav}
We also performed an ablation study to evaluate the efficacy of the developed \textit{Map Query Module} and \textit{Waypoint Selection Module} separately; no runtime changes occurred in this set of tests. 
In essence, we test how well human commands can be parsed to create a complete trajectory for the robot to navigate. 
This can be useful on its own in cases where a human operator needs to send a robot to explore an environment for specific features of interest. 
%Simulation settings are otherwise the same as previously. 
%are evaluated. As in Section \ref{experiment:simulation_result} we compare the routes to one selected by a person, who is constrained to the same number of waypoints and given the same feedback to pick waypoints and perform five trials for each route.

\begin{table}[!t]
    \vspace{6pt}
    \centering
    \caption{RMSE Values in Ablation Study Trajectories.}
    \vspace{-6pt}
    \begin{tabular}{ccc}
        \toprule
        & \multicolumn{1}{c}{Small House} & \multicolumn{1}{c}{Factory}\\
        %\cmidrule(lr){1-2}\cmidrule(lr){2-3} 
        %& RMSE & RMSE\\
        \midrule
        Route 1 & $0.905\pm0.622$ & $0.332\pm0.005$\\
        Route 2 & $1.298\pm1.239$ & $4.110\pm2.685$\\
        Route 3 & $1.443\pm0.570$ & $2.299\pm0.016$\\
        Average & $1.215\pm0.895$ & $2.247\pm2.187$\\
        \bottomrule
    \end{tabular}
    \vspace{-18pt}
    \label{tab:simulation_lang_nav}
\end{table}

RMSE scores are reported in Table~\ref{tab:simulation_lang_nav}. 
Overall these scores are higher compared to those reported in Table~\ref{tab:simulation_system_test} since all waypoints for the nominal trajectory were generated via our language processing method and not preset. 
%In the full system test, the waypoints before the obstacle are entered manually by a person, which results in a lower RMSE. 
The majority of the trajectories still follow the ground truth ones (Fig.~\ref{fig:simulated_lang_traj}).

\begin{figure*}
    \centering
    \subfloat[]{%[Route 1]{
        \frame{\includegraphics[trim=120 50 110 60, clip, width=0.23\textwidth]{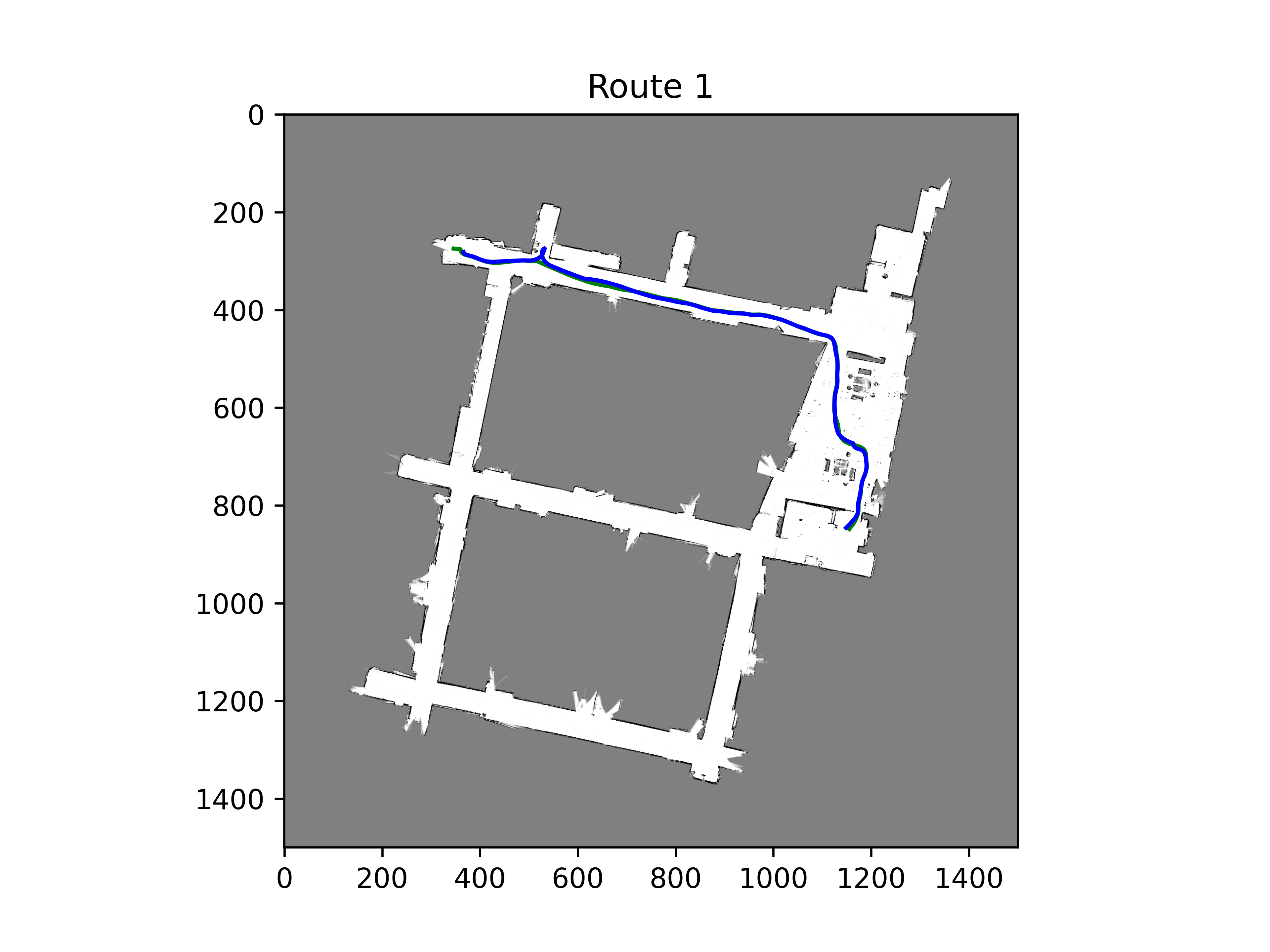}}
    }%
    \subfloat[]{%[Route 2]{
        \frame{\includegraphics[trim=120 50 110 60, clip, width=0.23\textwidth]{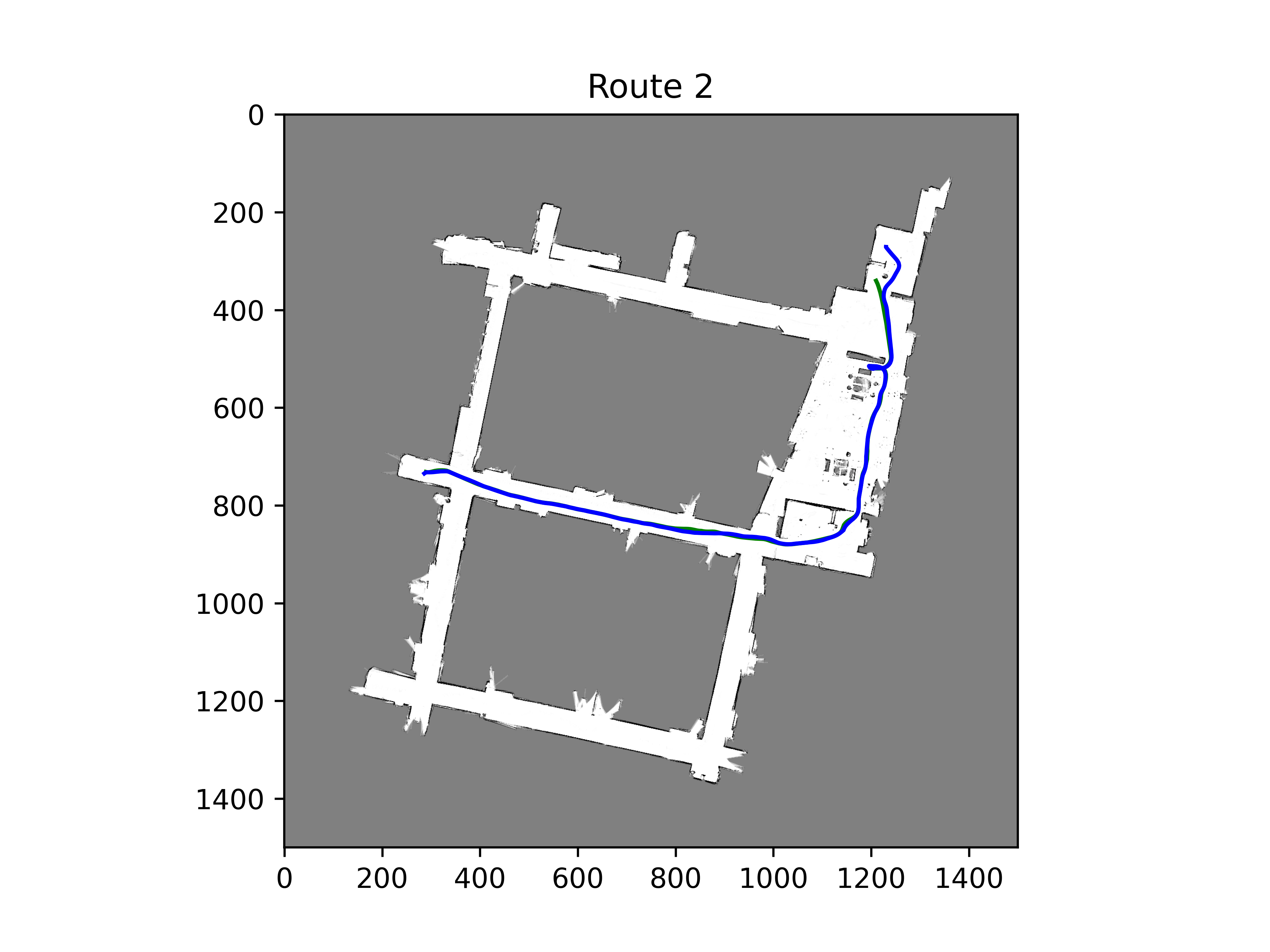}}
    }%
    \subfloat[]{%[Route 3]{
        \frame{\includegraphics[trim=120 50 110 60, clip, width=0.23\textwidth]{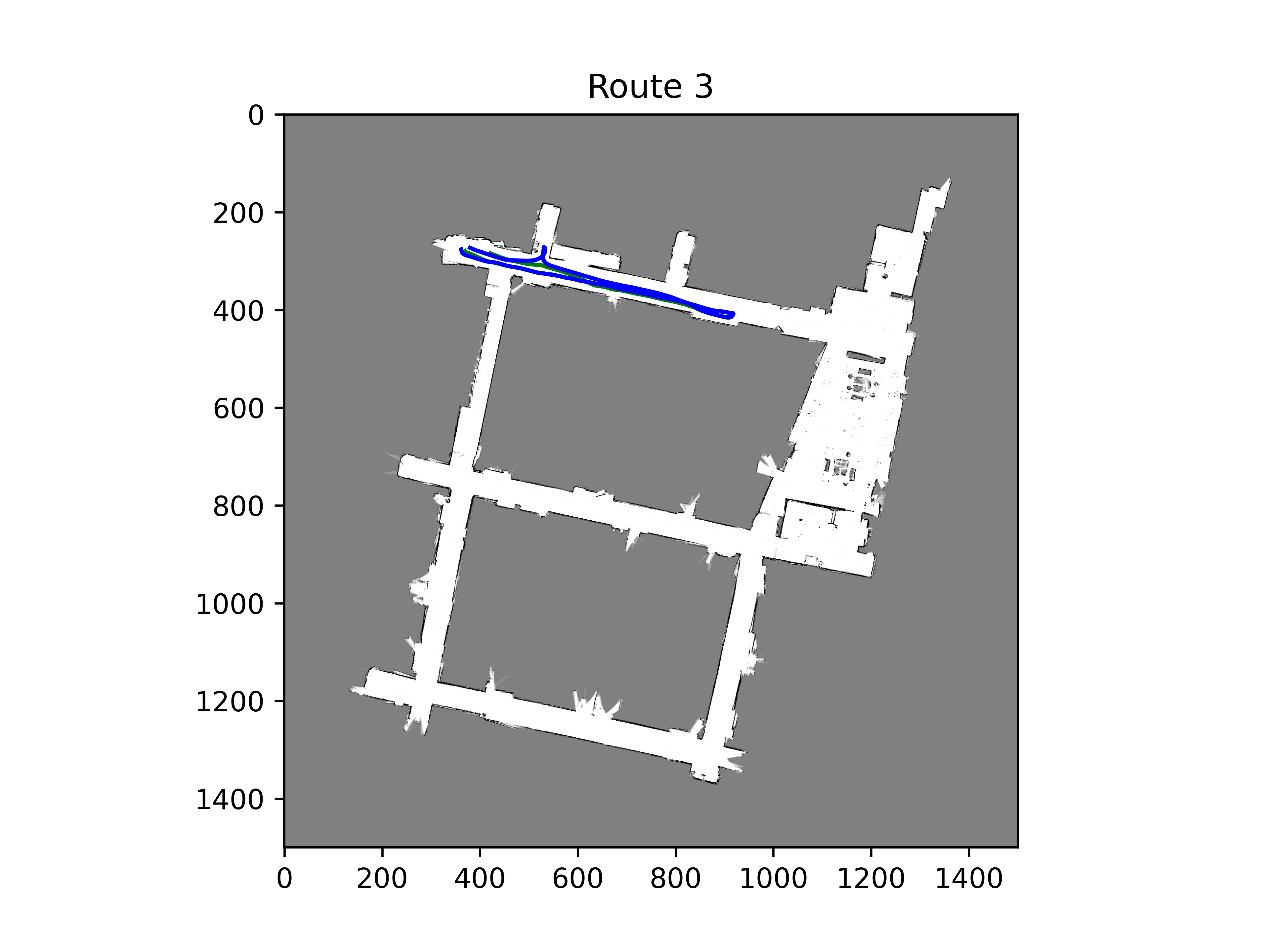}}
    }%
    \caption{Trajectories generated by our method (blue) overlaid to the ground truth (green) in the real-world experiments, for three distinctive routes.}%{Visualized results of our real-world experiments, with each trial drawn in blue and the ground truth trajectory drawn in green. We note that route 2 has a higher mean than the other routes since it consistently picks a final waypoint far from the ground truth endpoint.}%
    \label{fig:physical_system_traj}
    \vspace{-15pt}
\end{figure*}

\subsection{Real-World Experimentation Setup and Results}\label{experiment:physical_result}
We performed real-world experiments using a physical RosBot2.0 Pro from simulation, with an Orbec Astra RGBD camera, Slamtec RPLIDAR A3 LiDAR, and Intel Atom x5 Z8350 onboard computer. 
Navigation and the \textit{Query Decision Module} were run onboard. 
The \textit{Map Query Module} and \textit{Waypoint Selection Module} were run on a laptop with an Intel i5 CPU and no GPU, which communicated with the robot via Wi-Fi.
The operating environment was a floor of a building on our campus, comprising corridors with lab and classroom doors and a cafeteria that contained various types of furniture (couches, chairs, tables, etc.). 
%The RGB visualization is shown in Fig.~\ref{fig:physical_map}. 

%\input{figures/physical_map}
To construct the prior map for the real-world experiments, the robot was manually driven around the environment. 
Collected data in the real- world were significantly noisier than those in simulation, so we used cartographer~\cite{hess2016cartographer} to build higher quality maps and pose estimates for the semantic feature map creation. 
%In addition, the RGBD camera had significant trouble with determining the depth of geometries that were close to perpendicular to the image plane, making it nearly impossible to detect the ground plane, which can be seen in fig. \ref{fig:physical_map}. 
During execution time, we closed different doors between different hallways to act as our dynamic changes compared to the prior map, all of which were propped open during the initial mapping. 
We ran three different experiments and ran every experiment until we had collected five successful trials to ensure the RMSE scores could be calculated and was comparable across every experiments. A successful trial is defined by the proper triggering of the \textit{Query Decision Module}, if it fails to trigger or triggers where unintended we count that as a failure, and report the overall success rate.% \todo{write exactly what experiments were run, to get SR (not just five as per slack)}

\begin{figure}[!t]
    \centering
    \reflectbox{\includegraphics[angle=90,width=0.55\columnwidth]{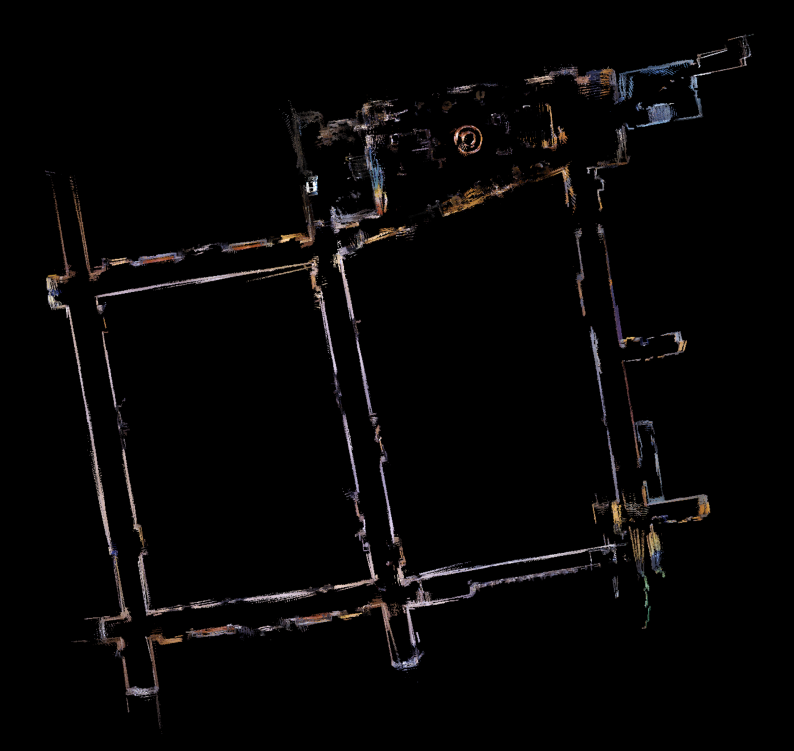}}
    \vspace{-6pt}
    \caption{RGB visualization of the prior map in real-world experiments.} %{The ground plane is not mapped, due to differences in the physical RGBD camera vs the simulated camera. Since this is collected in the real-world we have no ground truth rendering to compare to.}
    \label{fig:physical_map}
    \vspace{-18pt}
\end{figure}

We report the results of these experiments (RMSE and success rates) in Table~\ref{tab:physical_system_test}, and present the attained trajectories in Fig.~\ref{fig:physical_system_traj}. 
The RBG visualization of the prior map is depicted in Fig.~\ref{fig:physical_map}. 
Results demonstrate that the overall framework works in practical deployment as well. 
Routes that go through more variable parts of the map (Fig.~\ref{fig:physical_system_traj}(a)-(b)) demonstrate larger variability in RMSE and lower success rates as compared to navigating over less complex parts of the map (Fig.~\ref{fig:physical_system_traj}(c).) 
Compared to simulation results, in some cases, we noticed that onboard computational constraints resulted in messages being delivered out of order, which in turn led to underperformance.

\begin{table}[!t]
    \vspace{6pt}
    \centering
    \caption{Physical experiment system test results}
    \vspace{-6pt}
    \begin{tabular}{ccccc}
        \toprule
        & RMSE & SR\\
        \midrule
        Route 1 & $0.375\pm0.151$ & $0.63$\\
        Route 2 & $1.850\pm2.425$ & $0.83$\\
        Route 3 & $0.668\pm0.064$ & $1.00$\\
        Average & $0.964\pm1.541$ & $0.79$\\
        \bottomrule
    \end{tabular}
    \vspace{-6pt}
    \label{tab:physical_system_test}
\end{table}

\section{Conclusion}
This research focused on human-in-the-loop mobile robot navigation and developed an embodied AI system to enable bi-directional interactive communication between a human and a robot exploring an environment. 
To do so, we leveraged recent advances in foundational models to merge visual (from the robot) and language (from the human) information with conventional planning and control methods for autonomous robot navigation. 
Extensive testing in simulation and via physical experiments in the real world demonstrated the efficacy and robustness of our method in parsing human feedback in natural language and adapting the robot's trajectory accordingly both to adjust to dynamic changes in the environment as well as to create initial plans to explore an environment (in the ablation study). 
Results also revealed certain directions for improvement that constitute future work. 
One such direction is addressing sensor noise in pose estimation and the corresponding alignment with the semantic features map. 
In all, this work can serve as a foundation to extend deployment in more complex environments like orchards and construction sites.

%\textit{Query Decision Module} to decide when a robot should query a human for help. We showed how this feedback may be used to query a semantic feature map to produce a set of candidate waypoints with our \textit{Map Query Module} and to plan a new trajectory from these candidates with our \textit{Waypoint Selection Module}. 
%We performed a robust set of experiments to verify our method both in simulation and in real-world experiments. 
%We identified several shortcomings of current methods of semantic feature maps when applied to real-world settings. 
%Future works may explore how to automatically build semantic maps, how to build semantic maps with noisy pose data, or how to tightly integrate semantic information into robotic systems.

%%%%%%%%% REFERENCES
%{\small
\bibliographystyle{IEEEtran}
\bibliography{references}
%}

\vfill

\end{document}